\PassOptionsToPackage{hyperfootnotes=false}{hyperref}
\PassOptionsToPackage{obeyspaces}{url}
\documentclass[sigconf]{acmart-arxiv}

\makeatletter
\AtBeginDocument{\global\@ACM@balancefalse}
\renewcommand{\@authorfont}{\fontsize{10.5}{12.5}\selectfont}
\makeatother

\usepackage{booktabs} 
\usepackage{subcaption}

\graphicspath{{figure/}{figures/}}

\copyrightyear{2026}
\acmYear{2026}
\setcopyright{cc}
\setcctype{by}
\acmConference[MobiHoc '26]{The Twenty-seventh International Symposium on Theory, Algorithmic Foundations, and Protocol Design for Mobile Networks and Mobile Computing}{November 23--26, 2026}{Tokyo, Japan}
\acmBooktitle{The Twenty-seventh International Symposium on Theory, Algorithmic Foundations, and Protocol Design for Mobile Networks and Mobile Computing (MobiHoc '26), November 23--26, 2026, Tokyo, Japan}
\acmDOI{10.1145/3842721.3850098}
\acmISBN{979-8-4007-2975-1/2026/11}

\begin{document}
\title{Demo: Vision-Language Model-Guided Online Calibration of an Electromagnetic Digital Twin}

\newcommand{\afl}[1]{\texorpdfstring{\textsuperscript{#1}}{}}
\newcommand{\aand}{,\ }

\author{Zerui~Kang\afl{1}\aand Yishen~Lim\afl{1}\aand Zhouyou~Gu\afl{1}\aand
  Seungnyun~Kim\afl{1}\aand Seung-Woo~Ko\afl{2}\aand
  Tony~Q.S.~Quek\afl{1}\aand Jihong~Park\afl{1}}
\affiliation{%
  \institution{\afl{1}
               Singapore University of Technology and Design}
  \city{Singapore}
  \country{Singapore}
}
\affiliation{%
  \institution{\afl{2}Inha University}
  \city{Incheon}
  \country{Republic of Korea}
}
\email{zerui\_kang@mymail.sutd.edu.sg; swko@inha.ac.kr}
\email{{yishen\_lim,zhouyou\_gu,seungnyun\_kim,tonyquek,jihong\_park}@sutd.edu.sg}

\renewcommand{\shortauthors}{Z. Kang et al.}

\begin{abstract}
An electromagnetic (EM) digital twin gives mobile robots wireless situational awareness but depends on material conductivities that change with the environment. Online calibration faces initialization sensitivity and measurement travel costs. We demonstrate a vision-language model (VLM)-guided framework using a Unitree G1 robot and NVIDIA Sionna, with two VLM calls: material classification maps visible materials through ITU-R P.2040 to conductivity priors for Sionna’s gradient descent on accumulated received signal strength (RSS) measurements; waypoint planning selects the next measurement location online using residual RSS calibration error and image coverage. In a real indoor scenario, the framework achieves a normalized mean absolute conductivity error of $1.74\times10^{-4}$ within 20 m of travel; random initialization never converges, while random waypoints require over twice the travel.
\end{abstract}



\maketitle
\settopmatter{printacmref=true} %

\section{Introduction}
Electromagnetic (EM) digital twins model the radio propagation characteristics of a physical environment to provide mobile robots with situational awareness of the wireless environment. Their fidelity relies on estimates of environmental EM properties, which must be recalibrated from radio measurements as the environment changes. While existing methods estimate EM properties via differentiable ray tracing, e.g., NVIDIA Sionna~\cite{hoydis2023sionnart}, they typically operate \textit{offline} on pre-collected measurements. For example, Hoydis et al.~\cite{hoydis2024learning} learn material and antenna parameters from recorded indoor data, while Ai et al.~\cite{ai2026sparse} densify sparse measurements through a Bayesian channel map. To reduce the human effort of repeated measurement campaigns, mobile robots with onboard sensing offer a promising approach for \textit{online calibration}, where measurements are acquired progressively and travel-efficient selection becomes important. Our prior work~\cite{kang2026} uses a VLM to initialize material parameters and select measurement positions from a scene view, but these positions are fixed in advance and cannot adapt to evolving RSS calibration errors. Measurement positions should instead be selected from preceding observations to target poorly calibrated regions.

\begin{figure}[ht]
  \centering
  \captionsetup{skip=0.5pt}
  \includegraphics[width=\linewidth]{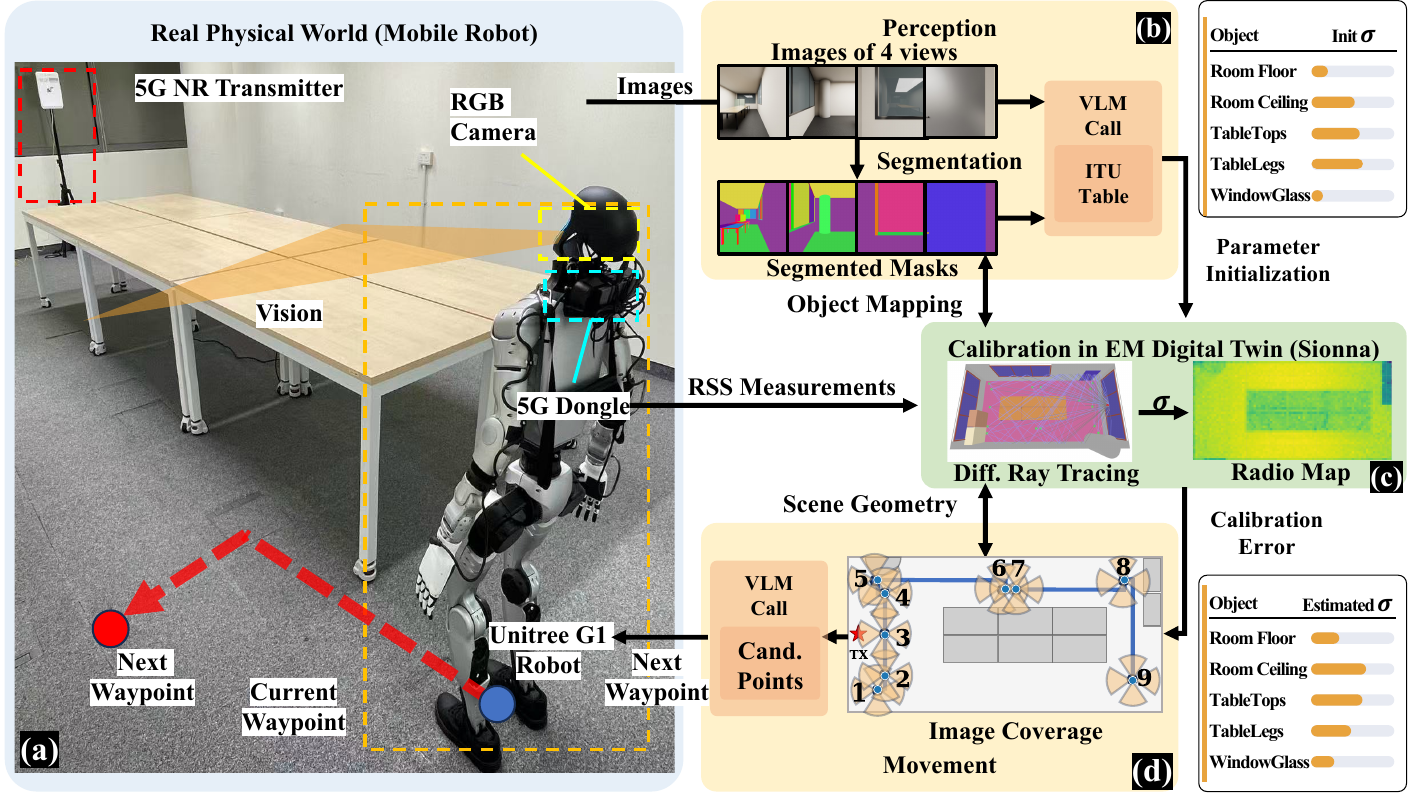}
  \Description{The proposed system consists of a VLM, a mobile robot, and a differentiable ray tracing engine. A waypoint-planning call selects the next waypoint from the calibration error and past image coverage, and a material-classification call maps the camera image to initial conductivity values for inverse ray tracing.}
  \caption{The proposed online calibration framework.}
  \label{fig:system}
\end{figure}\vspace{-4pt}

An aim of this paper is to propose a VLM-based online calibration framwork for an EM digital twin. The key idea is to couple two VLM calls in a perception-calibration-movement loop. Unlike~\cite{kang2026}, where the VLM initializes all parameters once from a complete view and fixes the measurement positions before calibration starts, both calls are made online in every round. First, since one viewpoint reveals only a few objects, the material parameters cannot be initialized once for the whole scene. A \emph{material-classification} call therefore initializes the objects newly observed at each waypoint from the camera images and their segmentation masks, which lets the VLM assign a material to each object rather than to the whole image. Priors close to the true values place the parameters where the RSS gradient points toward the true values, so every RSS measurement moves the parameters closer to the true values and the calibration converges within a few rounds. Second, since one measurement calibrates only the materials along its propagation paths, measuring where the nearby materials are already calibrated wastes travel. A \emph{waypoint-planning} call therefore selects the next waypoint from the RSS calibration error and the image coverage, so that each measurement reaches the materials still uncertain and each camera view covers the objects not yet observed, and the calibration converges with the least travel. Together, these two calls close the loop: visual observations initialize newly observed materials, RSS-based calibration errors guide the next waypoint, and newly collected images and RSS measurements feed into the next round.

\section{System Design}\label{sec:system}
We demonstrate the framework in the indoor scenario of Fig.~\ref{fig:system}(a) with a fixed 5G NR base station and a Unitree G1 humanoid robot carrying a 5G dongle and an RGB camera. A desktop with an NVIDIA RTX Pro 6000 GPU hosts the EM digital twin in Sionna and GPT-5.4 as the VLM. The dongle records RSS measurements from the fixed transmitter, while the Sionna configuration uses the same carrier frequency and antenna locations as the physical setup. The following three stages run once per round at each waypoint. For simplicity, we calibrate only the material conductivities.

\textbf{Perception:} sThe scene images in the four directions around the robot and the segmentation masks are provided to the \emph{*material-classification*} call, which labels the material of each segmented object, and each label is mapped through the ITU-R P.2040 database to a conductivity value~\cite{itur2040}, as shown in Fig.~\ref{fig:system}(b). The VLM outputs are conductivity priors for the visible objects and an image-coverage record of the objects observed so far, while objects outside all camera views retain their estimates from the previous round.

\textbf{Calibration:} Sionna in Fig.~\ref{fig:system}(c) refines the conductivities by gradient descent on the error between measured and simulated RSS, and reports a convergence indicator, namely the steadiness of the loss and the parameters over the last $50$ iterations. Since the VLM initializes the parameters close to the true values, gradient descent can quickly converge before the next measurement arrives, leaving small RSS-based calibration errors to the next round.

\textbf{Movement:} The RSS at a waypoint depends predominantly on the materials along the propagation paths from the base station to the robot, so a measurement can calibrate only those materials.
The \emph{waypoint-planning} call therefore takes the room layout, the RSS-based calibration error, and the image coverage, and picks the next waypoint whose propagation paths interact with the materials of still-uncertain conductivity and whose camera view covers objects not yet observed. Fig.~\ref{fig:system}(d) shows the waypoints selected online.
\section{Observations}
\begin{figure}[t]
  \centering
  \captionsetup{skip=0.5pt}
  \begin{subfigure}[t]{0.49\columnwidth}
    \centering
    \includegraphics[width=\linewidth]{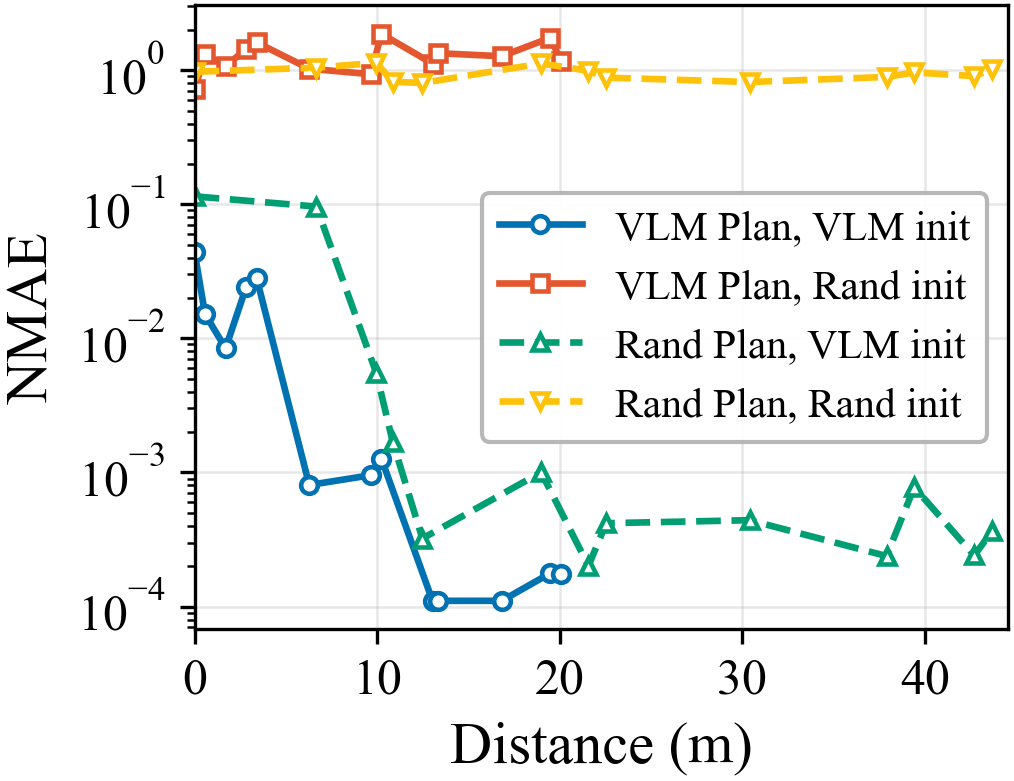}
    \caption{NMAE vs. traveled distance.}
    \label{fig:nmae}
  \end{subfigure}
  \hfill
  \begin{subfigure}[t]{0.48\columnwidth}
    \centering
    \includegraphics[width=0.7\linewidth]{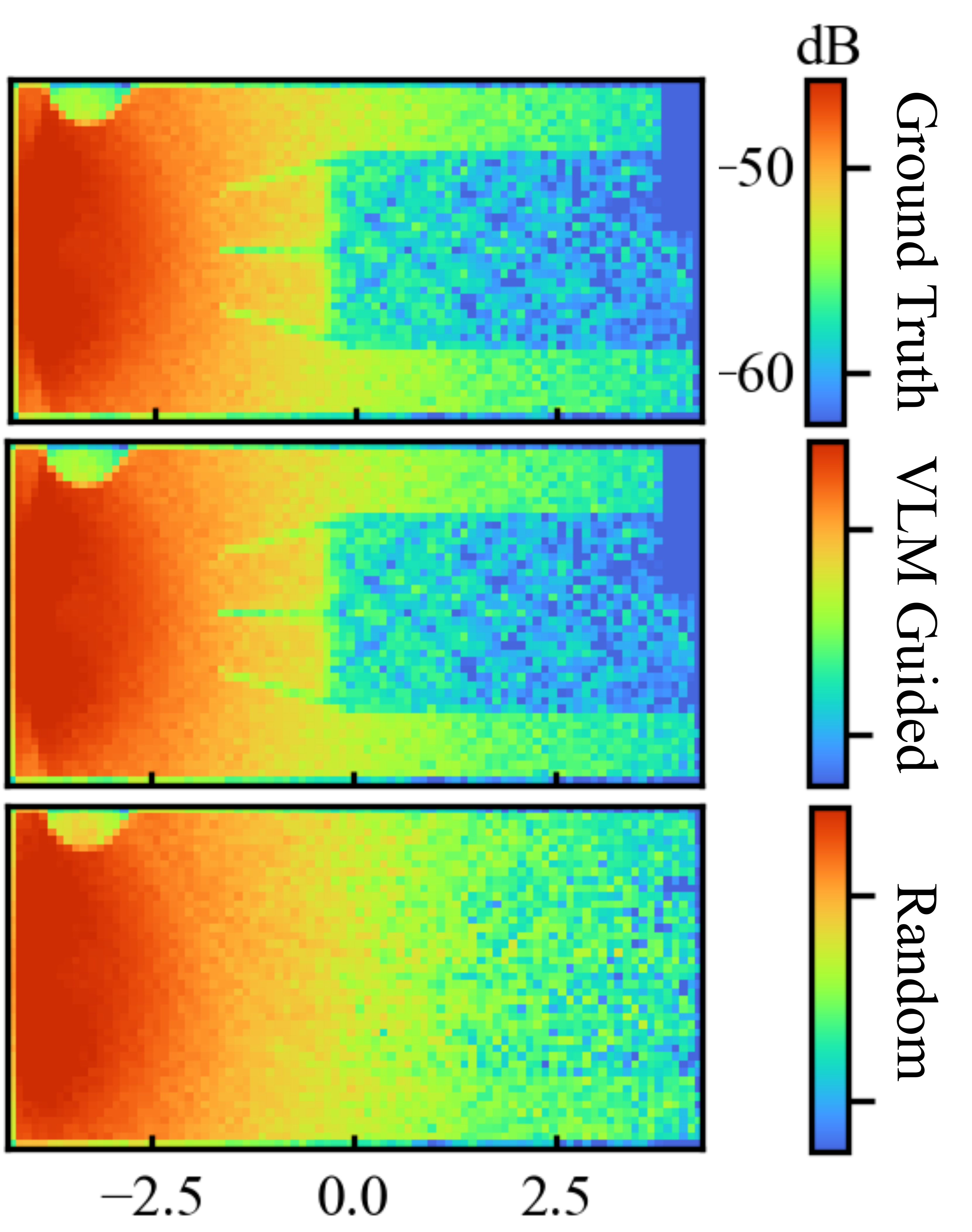}
    \caption{Calibrated radio maps.}
    \label{fig:plan_exp}
  \end{subfigure}
  \Description{Two subfigures. The left one plots NMAE against traveled distance for the four combinations of VLM or random waypoint planning and VLM or random initialization; only the VLM-initialized runs converge, and the VLM-planned run converges with the least travel. The right one compares the radio maps calibrated by the VLM-planned system and the random baseline.}
  \caption{Conductivity error and radio maps for the planning and initialization combinations.}
  \label{fig:results}
\end{figure}

We ablate each VLM call with a random baseline that draws initial conductivities at random or picks waypoints uniformly in the room. Fig.~\ref{fig:results}(a) plots the normalized mean absolute error (NMAE) of the estimated conductivity against traveled distance. For evaluation, the ground-truth conductivity of each material is predefined and used by Sionna to generate the radio map, while material parameters remain unknown to the digital twin throughout the calibration.

\textbf{Initialization decides convergence.} Both randomly initialized runs stay at an NMAE near unity (100\%) regardless of the planner, since the random values lie far from the true values, where gradient descent cannot update them in the correct direction. With the VLM priors, both planners converge, confirming that a material label places each conductivity where the RSS gradient points toward the nearby true value in the parameter space.

\textbf{Planning decides the travel needed to converge.} Given the priors, the waypoint-planning call reaches an NMAE of $1.74\times10^{-4}$ within 20.06~m, whereas the random planner needs 43.69~m to reach $3.66\times10^{-4}$ because it wastes travel where nearby materials are already calibrated. Fig.~\ref{fig:results}(b) shows the calibrated radio maps, where the proposed scheme attains a mean absolute error of $7\times10^{-3}$~dB against the ground truth, versus 2.027~dB for the random baseline.

\vspace{-2pt}
\section{Demonstration}
The demonstration runs the full pipeline in a photorealistic NVIDIA Isaac Sim replica of our meeting room, reproducing its layout, object placement, robot viewpoints, and measurement locations; real-world sensing noise and radio-hardware effects require separate validation. The EM digital twin and VLM run remotely in our laboratory. A live dashboard displays each round's material labels, robot movement to the selected waypoint, and calibrated radio map alongside the ground truth. Attendees can change object materials to trigger reclassification, recalibration, and map updates, or select measurement positions and compare performance with VLM planning. The setup requires network access and AC power for one laptop and one monitor. Video: \url{https://youtu.be/hwMRUpDA1PU}.
\vspace{-2pt}
\begin{acks}
This work was supported in part by the NVIDIA Academic Grant Program, in part by A$^\ast$STAR under its IAF-ICP (H25-MCP3438), in part by SUTD Kickstarter Initiative (SKI 2021 06 08), and in part by the National Research Foundation, Singapore and Infocomm Media Development Authority under its Communications and Connectivity Bridging Funding Initiative.
\end{acks}

\bibliographystyle{ACM-Reference-Format}
\bibliography{sigproc}

\end{document}